\documentclass{article}
\usepackage{spconf,amsmath,graphicx}
\usepackage{amssymb,amsfonts}
\usepackage{multirow}
\usepackage{textcomp}
\usepackage{xcolor}
\usepackage{pifont}
\usepackage{booktabs}
\usepackage[colorlinks]{hyperref}

\newcommand{\method}{SB-BEVFusion}

\title{SB-BEVFusion: Enhancing the Robustness against Sensor Malfunction and Corruptions}
\name{%
\begin{tabular}{@{}c@{}}
Markus Essl$^{1}$, Marta Moscati$^{1}$, Mubashir Noman$^{2}$, Muhammad Zaigham Zaheer$^{2}$, \\
Usman Naseem$^{3}$, Shah Nawaz$^{1}$, Markus Schedl$^{1,4}$
\end{tabular}
}

\address{%
$^{1}$Johannes Kepler University Linz, Austria,
$^{2}$MBZUAI, UAE, \\
$^{3}$Macquarie University, Sydney, Australia,
$^{4}$Linz Institute of Technology, Austria
}

\begin{document}
%
\maketitle
\begin{abstract}
Multimodal sensor fusion has demonstrated remarkable performance improvements over unimodal approaches in 3D object detection for autonomous vehicles. Typically, existing methods transform multimodal data from independent sensors, such as camera and LiDAR, into a unified bird's-eye view (BEV) representation for fusion. Although effective in ideal conditions, this strategy suffers from substantial performance deterioration when camera or LiDAR data are missing, corrupted, or noisy. To address this vulnerability, we develop a framework-agnostic fusion module for camera and LiDAR data that allows for handling cases when one of the two modalities is missing or corrupted.
To demonstrate the effectiveness of our module, we instantiate it in BEVFusion~\cite{liu2022bevfusion}, a well-established framework to combine camera and LiDAR data for 3D object detection. By means of quantitative experiments on the MultiCorrupt dataset, we demonstrate that our module achieves favorable performance improvements under scenarios of missing and corrupted modalities, substantially outperforming existing unified representation approaches across a wide range of sensor deterioration scenarios and reaching state-of-the-art performance in scenarios of corrupted modality due to extreme weather conditions and sensor failure.
\end{abstract}
\section{Introduction}
\vspace{-0.7em}
Perception in self-driving cars is predominantly dependent on two complementary sensors: \textit{camera} and \textit{LiDAR}~\cite{yan2023cross}. The former sensor is responsible for rich visual appearance, while the latter offers the precise geometry necessary for accurate detection of objects~\cite{feng2020deep}. To exploit this complementary information from camera and LiDAR, techniques for Bird's-Eye View (BEV) such as BEVFusion \cite{liang_bevfusion_2022, liu2022bevfusion} propose to fuse these two modalities, assuming that both are present and of high quality. However, real-world scenarios are not ideal and may lead to corrupted data, resulting in substantial performance deterioration of such systems~\cite{liaqat2025multimodal,yu_benchmarking_2023,park_resilient_2025}. These scenarios include situations like adverse weather conditions affecting the visibility from the camera, such as fog, rain, and malfunctions affecting the LiDAR system, such as beam reduction and occlusion~\cite{xie2023benchmarking,dong2023benchmarking,beemelmanns2024multicorrupt}. Furthermore, spatial and temporal misalignment between the two modalities may cause considerable deterioration of the systems. 
Addressing these situations is an emerging field of research. For example, a recent work ~\cite{beemelmanns2024multicorrupt} has mainly focused on curating a benchmark to evaluate the state-of-the-art (SOTA) multimodal 3D detectors on their robustness against sensor failures or corruptions.

To this end, we set out to develop a fusion technique capable of robustly handling corrupted or missing modalities, and at the same time, it can be seamlessly integrated as a fusion module in existing architectures. 
Specifically, we introduce a single-branch BEV detector (SB-BEVFusion) that allows the model to maintain good performance 
when camera or LiDAR data is missing. When both modalities are clean and present, the BEV features 
are combined by means of a pipeline relying on unweighted averaging, max pooling, cross-attention, and progressive modality decay. When one modality is absent or corrupted, our proposed architecture passes the available BEV representations without fusion. The novelty of our model relies on the fact that the downstream network and the detection head are shared across both modalities (camera (C) and LiDAR (L)) and trained on a shuffled mixture of L+C/L/C. This allows our model to handle both unimodal and multimodal scenarios in an effective way, therefore tackling the case of missing modality effectively.
In summary, our main contributions are as follows.

\begin{itemize}
  \item We investigate various fusion strategies to combine the multimodal BEV feature representations, focusing on their robustness to missing and corrupted modality.

  \item We introduce a module that is designed to explicitly operate under L+C/L/C scenarios. 

  \item We conduct extensive experiments on the MultiCorrupt dataset and empirically demonstrate that our proposed module 
  demonstrates improved robustness towards the scenarios of missing and corrupted modality.

\end{itemize}

\begin{figure*}[t]
  \centering
  \includegraphics[width=0.78\linewidth]{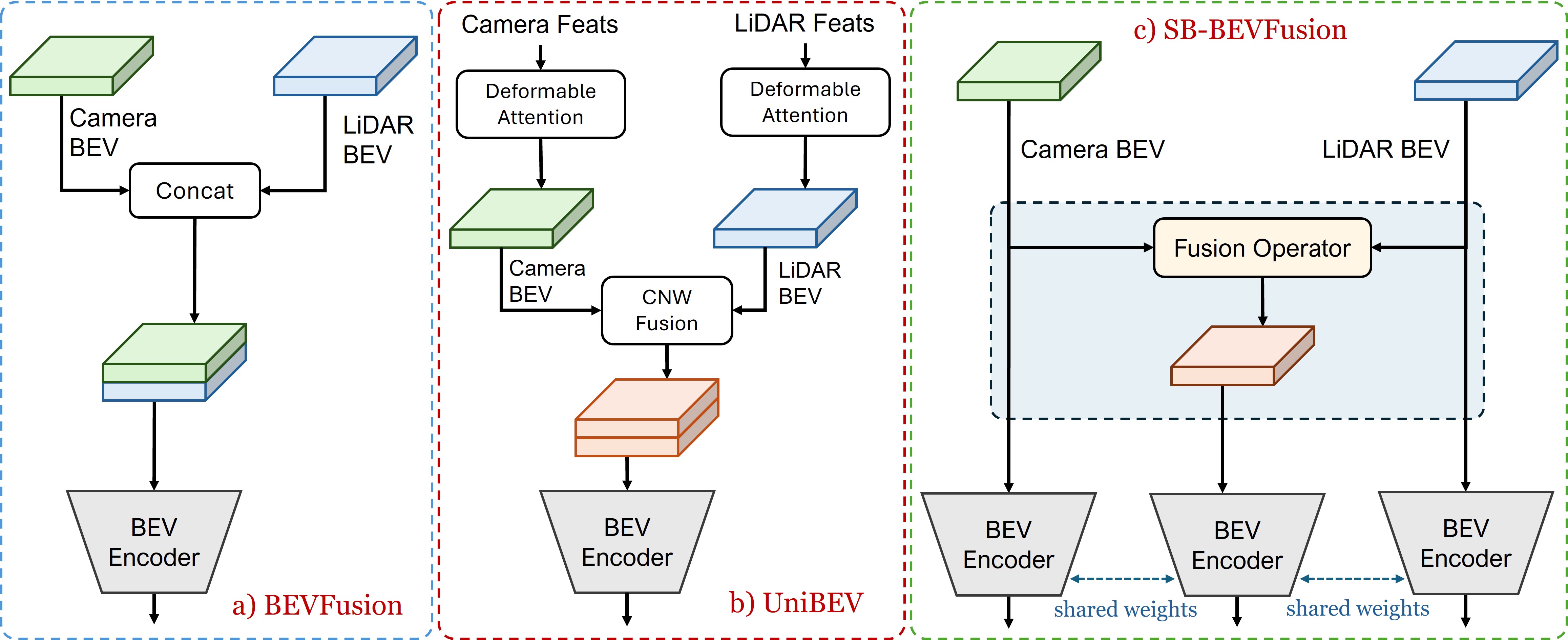}
  \vspace{-0.7em}
  \caption{Comparison between proposed \textbf{SB-BEVFusion}, BEVFusion~\cite{liu2022bevfusion}, and UniBEV~\cite{wang_unibev_2024}. a) Camera and LiDAR data is processed by modality-specific encoders to obtain features which are transformed to camera and LiDAR BEV representations. Afterwards, these BEV features are concatenated and processed by BEV encoder. b) UniBEV uses deformable attention before transformation to BEV space. It utilizes Channel Normalized Weights (CNW) fusion module to combine the BEV representations and feed them to BEV encoder. c) Proposed SB-BEVFusion does not add complexity to the architecture and combines the BEV representations using a fusion operator $\mathcal{F}$ when both modalities are available or acts as an identity when one of them is absent. A shared BEV encoder processes the BEV features $\mathbf{F}_{\text{in}}$ in all modes to obtain 3D detections.}
  \label{fig:SB-BEVFusion}
\end{figure*}

\begin{figure*}[t]
  \centering
  \includegraphics[width=0.78\linewidth]{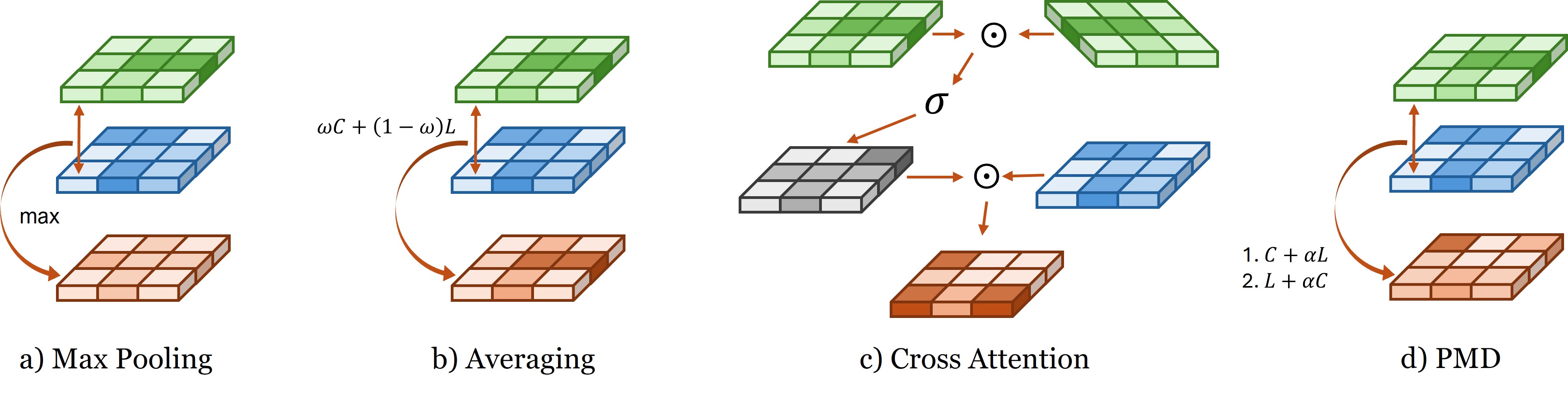}
  \vspace{-0.7em}
  \caption{Illustration of various fusion operations to combine the BEV feature representations utilized in this study.}
  \label{fig:fusion_ops}
\end{figure*}

\section{Methodology}
\label{sec:method}
\vspace{-0.7em}

\noindent \textbf{Setup and Notation.}
We adopt the pipeline of BEVFusion \cite{liu2022bevfusion}, depicted in Fig.~\ref{fig:SB-BEVFusion}, as underlying framework to apply our fusion module. As in BEVFusion, we first extract feature representations utilizing modality-specific encoders for the LiDAR point cloud $\mathbf{P}$ and for the  multi-view camera images $\mathbf{I}=\{I_k\}_{k=1}^{K}$, where $K$ represents the number of camera views.
These 
representations are then projected to a unified BEV space preserving the semantic and geometric information. The resulting representations in BEV space are given by
\begin{equation}
\tilde{\mathbf{F}}_{\text{cam}} \in \mathbb{R}^{H \times W \times C_{\text{cam}}}, \qquad
\tilde{\mathbf{F}}_{\text{lid}} \in \mathbb{R}^{H \times W \times C_{\text{lid}}}
\end{equation}
%
where $H, W, C$ represent the height, width, and number of channels, respectively.
BEVFusion combines $\tilde{\mathbf{F}}_{\text{cam}}$ and $\tilde{\mathbf{F}}_{\text{lid}}$, 
with simple concatenation $\mathcal{F}$, obtaining the multimodal representation which is fed to a convolutional BEV encoder 
$\mathcal{E}$, designed to adjust the local misalignments that can occur due to inaccurate depth.
\begin{equation}
\mathbf{F}_{\text{fused}} = \mathcal{F}\left(\tilde{\mathbf{F}}_{\text{lid}}, \tilde{\mathbf{F}}_{\text{cam}}\right), \qquad
\hat{\mathbf{y}} = \mathcal{E}\left(\mathbf{F}_{\text{fused}}\right)
\end{equation}
%
%
It is well-known that the fusion operation is critical for effective combination of multimodal information~\cite{arevalo2017gated,saeed2022fusion}. 
This motivated us to investigate various fusion schemes and analyze their impact on improving model performance under different sensor corruptions. 
Moreover, our concern for the model robustness against sensor failure further encouraged us to introduce the \method, which is based on effective training strategy and fusion operators and does not assume both modalities to be present. 

To this end, we first equalize channel widths such that $C_{\text{cam}}=C_{\text{lid}}\triangleq C$,
yielding BEV tensors of same dimensions. This allows us to apply fusion $\mathcal{F}$ when both modalities are present and only pass the features of the available modality to the BEV encoder otherwise, as illustrated in Fig. \ref{fig:SB-BEVFusion}.
\begin{equation}
\mathbf{F}_{\text{in}} =
\begin{cases}
\mathcal{F}\left(\tilde{\mathbf{F}}_{\text{lid}}, \tilde{\mathbf{F}}_{\text{cam}}\right), & a_{\text{lid}} \land a_{\text{cam}},\\
\tilde{\mathbf{F}}_{\text{lid}}, & a_{\text{lid}},\\
\tilde{\mathbf{F}}_{\text{cam}}, & a_{\text{cam}}.
\end{cases}
\qquad
\hat{\mathbf{y}} = \mathcal{E}\left(\mathbf{F}_{\text{in}}\right)
\end{equation}
%

\noindent \textbf{Fusion Operators.}
The fusion operator $\mathcal{F}$ is responsible for mapping the same-sized multimodal BEV representations into a combined single BEV tensor, given by 
$\mathbf{F}_{\text{in}}=\mathcal{F}(\tilde{\mathbf{F}}_{\text{lid}},\tilde{\mathbf{F}}_{\text{cam}})\in\mathbb{R}^{H\times W\times C}$. The fusion operator is applied when both modalities are present and is replaced by the identity function when one modality is missing. The fusion operators investigated in our approach are depicted in Fig. \ref{fig:fusion_ops}.

\begin{enumerate}
  \item \textbf{Averaging: } We first combine the camera and LiDAR BEV representations by using unweighted average fusion. This is given by: 
  \vspace{-2mm}
  \begin{equation}
\mathbf{F}_{\text{in}} = w\,\tilde{\mathbf{F}}_{\text{cam}} + (1-w)\,\tilde{\mathbf{F}}_{\text{lid}}, \qquad w = 0.5
\end{equation}
  \vspace{-4mm}  

  \item \textbf{Max-Pooling: } 
  We apply the max-pooling operation on the two BEV representations and retain 
  features given by:
  \vspace{-2mm} 
  \begin{equation}
\mathbf{F}_{\text{in}} =
\max\!\left(\tilde{\mathbf{F}}_{\text{lid}},\, \tilde{\mathbf{F}}_{\text{cam}}\right)
\quad \text{(element-wise over } H \times W \times C \text{)}
\end{equation}

  \item \textbf{Cross-Attention: } Motivated by 
  the performance improvement on other multimodal domains~\cite{liang2021attention, SHEN2024_ICAFusion}, we utilize cross-attention to merge the camera and LiDAR BEV features to obtain the refined fused representations. Mathematically, this operation is given by:
  \vspace{-2mm} 
  \begin{equation}
\mathbf{F}_{\text{in}}
=
\tilde{\mathbf{F}}_{\text{lid}}
+
\gamma\, W_o\,
\mathrm{Attn}\!\left(
W_q \tilde{\mathbf{F}}_{\text{lid}},
W_k \tilde{\mathbf{F}}_{\text{cam}},
W_v \tilde{\mathbf{F}}_{\text{cam}}
\right)
\end{equation}
    
    Here, $\mathrm{Attn}(\cdot)$ is standard multi-head attention; $W_{\{\cdot\}}$ are $1{\times}1$ projections;
    we flatten $C\!\times\!H\!\times\!W$ to $HW$ tokens inside $\mathrm{Attn}$ and reshape back afterwards;
    $\gamma=\sigma(\theta)$ is a scalar gate.

  \item \textbf{Progressive Modality Decay (PMD): } To encourage the model to adapt to the missing-modality case, we combine the BEV representations in a modality-decay fashion. In this case, one modality is kept constant while second modality is multiplied by a decaying factor $\alpha$ and added to the first modality. 
  For each training example, an anchor modality is chosen. The fused BEV features are obtained as:
  \vspace{-2mm} 
\begin{equation}
\mathbf{F}_{\text{in}} =
\tilde{\mathbf{F}}_{\text{anchor}} + \alpha\,\tilde{\mathbf{F}}_{\text{other}}
\end{equation}
  \vspace{-2mm} 
  where $\alpha$ decays from $1$ (start of training) to $0$ (end).
\end{enumerate}

\noindent Among the investigated fusion operators, \textit{Averaging} demonstrates superior performance, as detailed in Table \ref{tab:nds-sota-comparison-horizontal}. Therefore, this is a default choice in our experiments, unless specifically stated otherwise.

\noindent \textbf{Training Schedules.}
Following prior work~\cite{tschannen2023clippo,liaqat2025chameleon,ganhor2025single,ganhor2024multimodal,saeed2024modality}, we train a single downstream path to operate under three regimes: L{+}C, L, and C. 
For the average, max-pooling and cross-attention fusion strategies, 
we enumerate all three regimes per sample; for PMD, we enumerate two anchored passes (LiDAR-anchored and camera-anchored). The expanded epoch is globally shuffled so each mini-batch mixes regimes. Concrete epoch counts are given in Sec.~\ref{sec:experimental_setup}.


\noindent \textbf{Inference.}
During inference time, we utilize the fusion operator only when both modalities are available. If one modality is missing, the fusion operator becomes the identity, and the model runs with only one modality as input, with no other structural change. This scheme enables the model to handle missing-modality scenarios and enhance its robustness.
\vspace{-0.7em}


\begin{table*}[t]
\caption{Comprehensive NDS comparison of SB-BEVFusion variants, fusion strategies, and SOTA under corruptions. Abbreviations: BEVF (BEVFusion), UniBEV (UniBEV\_cnw), BEVFusion-D (Decay), BEVFusion-Avg (Avg), BEVFusion-CA (Cross-Attention), BEVFusion-MP (Max-Pool).}
\centering
\vspace{3mm}
\resizebox{0.85\linewidth}{!}{
\begin{tabular}{llcc|cccc}
\toprule
\textbf{Corruption} & \textbf{Sev.} & \textbf{BEVF} & \textbf{UniBEV} & \textbf{SB-BEVFusion-D} & \textbf{SB-BEVFusion-Avg} & \textbf{SB-BEVFusion-CA} & \textbf{SB-BEVFusion-MP} \\
\midrule
\multirow{3}{*}{Beam Reducing} & $s1$ & 0.6338 & 0.6181 & 0.5764 & 0.6219 & 0.6380 & 0.6365 \\
 & $s2$ & 0.4818 & 0.5061 & 0.4213 & 0.4821 & 0.4840 & 0.4954 \\
 & $s3$ & 0.3052 & 0.3981 & 0.2774 & 0.3192 & 0.3181 & 0.3303 \\
\midrule
\multirow{3}{*}{Camera Fog} & $s1$ & 0.6476 & 0.6363 & 0.6064 & 0.6464 & 0.6542 & 0.6520 \\
 & $s2$ & 0.5978 & 0.5931 & 0.5600 & 0.5999 & 0.6072 & 0.6043 \\
 & $s3$ & 0.3453 & 0.3648 & 0.3466 & 0.3565 & 0.3740 & 0.3615 \\
\midrule
\multirow{3}{*}{Motion Blur} & $s1$ & 0.6687 & 0.6348 & 0.6287 & 0.6629 & 0.6789 & 0.6769 \\
 & $s2$ & 0.5865 & 0.5344 & 0.5447 & 0.5912 & 0.6060 & 0.6033 \\
 & $s3$ & 0.4991 & 0.4427 & 0.4634 & 0.5199 & 0.5250 & 0.5246 \\
\midrule
\multirow{3}{*}{Spatial Misalignment} & $s1$ & 0.5836 & 0.5794 & 0.5293 & 0.5917 & 0.5820 & 0.5896 \\
 & $s2$ & 0.4937 & 0.5065 & 0.4368 & 0.5110 & 0.4891 & 0.5032 \\
 & $s3$ & 0.4243 & 0.4492 & 0.3682 & 0.4408 & 0.4215 & 0.4351 \\
\midrule
\multirow{3}{*}{Temporal Misalignment} & $s1$ & 0.6276 & 0.6122 & 0.5920 & 0.6254 & 0.6348 & 0.6351 \\
 & $s2$ & 0.5394 & 0.5292 & 0.5091 & 0.5405 & 0.5480 & 0.5480 \\
 & $s3$ & 0.4676 & 0.4646 & 0.4456 & 0.4705 & 0.4785 & 0.4768 \\
\midrule
\multicolumn{2}{l}{\textbf{Mean Resistance Ability}}  &  0.7490  &  \underline{0.7656}  &  0.7313  &  \textbf{0.7683 } &  0.7537  &  0.7592  \\
\bottomrule
\end{tabular}
}
\label{tab:nds-sota-comparison-horizontal}
\end{table*}

\section{Experiments}
\label{sec:experimental_setup}
\vspace{-0.7em}
\noindent \textbf{Dataset.}
We evaluated our proposed strategies on the nuScenes dataset~\cite{caesar2020nuscenes}.
This is a large-scale multimodal benchmark dataset for autonomous driving, providing full 360° coverage, including six cameras, one $32$-beam LiDAR, and five radars; this study focuses on LiDAR and camera data streams, and we therefore discard the radar data.
The dataset comprises $1,000$ scenes, each $20$ seconds long
. These scenes were captured in Boston (USA) and Singapore, and provide samples of varied driving conditions, times of day (day/night), and weather conditions (e.g., clear, rain). The dataset also provides rich 3D bounding box annotations for $23$ object classes; the $10$ common classes from the official nuScenes detection challenge are used for 3D object detection.
Model training and evaluation use the official nuScenes train and val splits with $850$ and $150$ scenes, respectively, for consistency with established benchmarks. 
We applied standard pre-processing steps, aligned with common BEVFusion methodologies \cite{liu2022bevfusion}. Camera images are resized to a fixed resolution and normalized. For LiDAR data, points from the previous $10$ sweeps are accumulated to form a denser point cloud for the current frame. These accumulated points are transformed into the current ego-vehicle's reference frame, accounting for vehicle motion and sensor extrinsic calibration.
Moreover, we use the MultiCorrupt~\cite{beemelmanns2024multicorrupt} dataset as a benchmark to evaluate our method against various corruption categories and severity levels.
The dataset was intentionally corrupted to simulate challenging real-world environmental conditions.
Corruption severity is defined across three levels, following the MultiCorrupt benchmark: Level 1 introduces minor perturbations to simulate mild environmental noise, Level 2 applies medium-strength distortions, and Level 3 imposes severe corruption to replicate extreme failure scenarios.

\begin{table*}[t]
\caption{Overall performance on nuScenes validation set with multiple missing sensor configurations. Best results are bold; second best are underlined.
}
\vspace{3mm}
\centering
\resizebox{0.85\linewidth}{!}{
\begin{tabular}{l|ccc|ccc}
\toprule
\multirow{2}{*}{\textbf{Modality Availability}} & \multicolumn{3}{c|}{\textbf{mAP $\uparrow$}} & \multicolumn{3}{c}{\textbf{NDS $\uparrow$}} \\
\cmidrule(lr){2-4} \cmidrule(lr){4-7}
& \textbf{SB-BEVFusion (Avg)} & \textbf{BEVFusion} & \textbf{UniBEV} & \textbf{SB-BEVFusion (Avg)}  & \textbf{BEVFusion} & \textbf{UniBEV} \\
\midrule
Both Modalities & \textbf{0.6737} & \underline{0.6691} & 0.642 & \underline{0.6970}  & \textbf{0.7033} & 0.685 \\
Camera Only     & \underline{0.2002} & 0.0109 & \textbf{0.35 }& 0.2440  & 0.1074 & 0.424  \\
LiDAR Only      & \textbf{0.6448} & 0.5639 & \underline{0.582} & \textbf{0.6959}  & 0.5361 & \underline{0.653}   \\
\bottomrule
\end{tabular}
}
\label{tab:summary-map-nds}
\end{table*}

\begin{figure*}[t]
  \centering
  \includegraphics[width=0.78\linewidth]{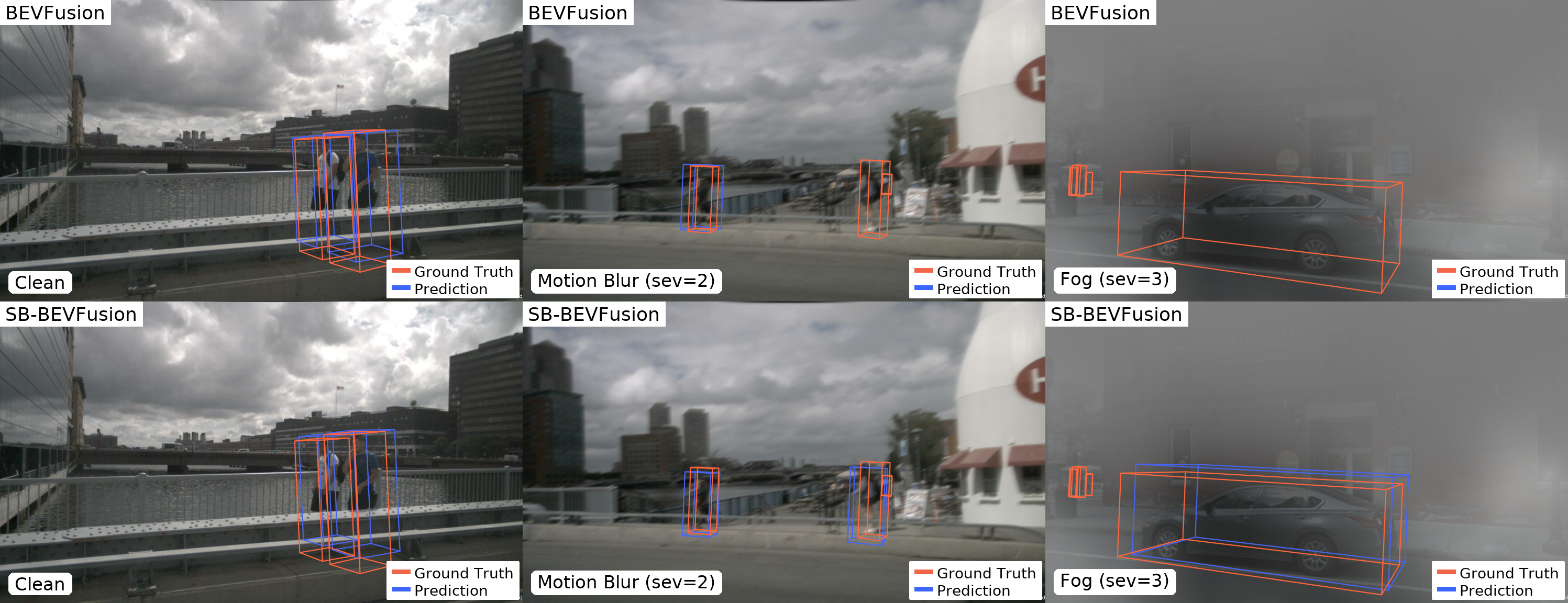}
  \vspace{-0.7em}
  \caption{\textbf{Qualitative comparison of BEVFusion (top) vs.\ SB-BEVFusion (bottom).} Panels show Clean, Motion Blur (severity 2), and Fog (severity 3). Under corruptions, SB-BEVFusion remains robust compared to BEVFusion’s predictions, matching quantitative trends. We observe that BEVFusion detection performance deteriorates notably with increasing severity level, i.e. Motion Blur and Fog. }
  \label{fig:qual-SB-BEVFusion-mb-fog}
\end{figure*}

\noindent \textbf{Metrics and Protocol.}
We report the standard nuScenes detection metrics: mean Average Precision (mAP) and nuScenes Detection Score (NDS). Unless stated otherwise, all results are on the official \texttt{val} split. We evaluate three modality-availability regimes: \textbf{L{+}C} (both sensors), \textbf{L} (LiDAR only), and \textbf{C} (camera only). 
Moreover, we evaluate robustness with MultiCorrupt \cite{beemelmanns2024multicorrupt} on nuScenes \texttt{val}, using $5$ corruption families including LiDAR beam reduction, spatial misalignment, temporal misalignment, fog, and motion blur.
Unless stated otherwise, corruptions are applied to raw sensor streams prior to BEV projection, with three severities $s1$–$s3$.
Besides the standard NDS, we report the \emph{mean Resistance Ability (mRA)}, defined as the average over corruption families and severities of the ratio between a metric \(\mathcal{M}\) measured under corruption and its value on clean data; in our experiments, \(\mathcal{M}=\mathrm{NDS}\):
\vspace{-0.7em}
\[
\mathrm{mRA}=\frac{1}{3N}\sum_{c=1}^{N}\sum_{s=1}^{3}\frac{\mathcal{M}_{c,s}}{\mathcal{M}_{\mathrm{clean}}}.
\]

\noindent \textbf{Implementation Details.}
We match the official BEVFusion configuration and initialize from a parity checkpoint. The only deviations from the public config are: (i) we set the camera BEV projection width to match the LiDAR BEV channels so that $C_{\text{cam}}=C_{\text{lid}}(\!\triangleq C)$, we also apply this change when training the BEVFusion baseline, and use that checkpoint for all ablations; (ii) for \method we replace the two-branch concat+conv with our single-branch module defined in Sec.~\ref{sec:method} while preserving the BEV tensor interface; and (iii) we train with availability mixing by enumerating \{L{+}C, L, C\} per sample and shuffling the expanded epoch. Upstream encoders are frozen to isolate downstream effects. We train the 3-regime operators for $3$ epochs and PMD for $4$ epochs over their expanded datasets.

\section{Results and Discussion}
\vspace{-0.7em}
\noindent \textbf{Robustness Against Corrupted Sensor Modalities.}
Table \ref{tab:nds-sota-comparison-horizontal} provides an extensive performance evaluation of our method alongside other SOTA strategies  under various types of  corruptions.
The experimental results demonstrate that all methods experience performance deterioration as the severity of corruptions increases, but the extent of degradation varies across approaches. 
For beam reduction and motion blur, the attention-based fusion methods (e.g., SB-BEVFusion-CA and SB-BEVFusion-MP) maintain relatively stronger robustness compared to other variants. Under camera fog, most methods achieve competitive results, with SB-BEVFusion-CA slightly outperforming others. For spatial and temporal misalignments, SB-BEVFusion-CA consistently shows superior resilience across different severities. When averaged across all corruptions, SB-BEVFusion-Avg achieves the highest mean resistance ability ($0.7683$), closely followed by UniBEV ($0.7656$), SB-BEVFusion-MP ($0.7592$). 
These findings highlight that different fusion strategies have varying sensitivity to corruption types, with attention- and averaging-based fusion methods showing the most stable performance.

\noindent \textbf{Robustness Against Missing Sensor Modalities.}
Although corruption affects the overall quality of a sensor, some information may still be retained, which the network may use to perform the detection task.
In this section, we consider a more extreme scenario, i.e., the setting in which a sensor is completely removed. This may be considered a corner case where no data regarding the corrupted sensor is available to the network.
Table \ref{tab:summary-map-nds} presents a comprehensive evaluation of our method alongside other SOTA strategies under various conditions of missing sensor data.
The results demonstrate a clear trade-off between peak performance and robustness across the evaluated models. When both camera and LiDAR data are available, our method, SB-BEVFusion, achieves the highest detection precision (mAP of $0.6737$). 
Under sensor failure scenarios, a notab.le performance contrast emerges: UniBEV proves substantially more robust in camera-only settings (mAP $0.35$), outperforming other approaches. 
We conjecture that such favorable performance is obtained due to the utilization of deformable attention before transforming the multiview camera features into BEV representations. 
Conversely, in the LiDAR-only condition, our method excels (mAP $0.6448$), demonstrating superior robustness. These findings indicate that while BEVFusion performs well under ideal sensing conditions, our approach maintains competitive performance in full sensor settings while exhibiting superior LiDAR-only robustness. In contrast, UniBEV shows distinctive resilience to camera-only operation.

\noindent \textbf{Qualitative Analysis.}
Fig.~\ref{fig:qual-SB-BEVFusion-mb-fog} presents a qualitative comparison between BEVFusion (top row of each panel) and our SB-BEVFusion (bottom row of each panel) under different corruption scenarios. On clean data, both methods provide accurate predictions closely aligned with the ground truth. However, under corruptions such as motion blur (severity level 2) and fog (severity level 3), differences in performance become evident. 
We observe from Fig.~\ref{fig:qual-SB-BEVFusion-mb-fog}, even notably large objects are not detected by BEVFusion when fog severity level is highest. 
In general, BEVFusion exhibits deteriorated detection performance, with predicted bounding boxes deviating more noticeably from the ground truth. Alternatively, SB-BEVFusion produces more accurate predictions under the same conditions, maintaining better alignment with the ground truth boxes despite the presence of corruptions. These qualitative results confirm the quantitative findings, highlighting that SB-BEVFusion is more resilient to corruptions, especially under severe conditions.
\vspace{-0.7em}


\section{Conclusion}
\vspace{-0.7em}
In this work, we proposed SB-BEVFusion method that improves the robustness of the model against sensor corruption and failure. The paper investigated several fusion operators to effectively combine multimodal BEV representations and demonstrated that unweighted averaging provides superior overall performance compared to other fusion operators. 
Extensive experimentation on MultiCorrupt dataset
demonstrated that the concatenation operation for BEV feature fusion is suboptimal and requires careful considerations for selecting the effective fusion operations depending on the corruption type. 
In addition, utilizing the shuffled mixture of L+C/L/C scenarios for model training improves its robustness against modality failure.
One of the limitations of the proposed work is that, although  SB-BEVFusion would allow fusing other modalities (e.g., radar) after converting them to BEV, we only tested it on camera and LiDAR data. 
Our potential future direction is to further investigate and improve performance of transformer based fusion methods.

\section{Acknowledgments}
   \vspace{-0.7em}
This research was funded in whole or in part by the Austrian Science Fund (FWF): Cluster of Excellence \href{https://www.bilateral-ai.net/home}{\textcolor{blue}{\textit{Bilateral Artificial Intelligence}}} (\url{https://doi.org/10.55776/COE12}), the doc.funds.connect project \href{https://dfc.hcai.at/}{\textcolor{blue}{\textit{Human-Centered Artificial Intelligence}}} (\url{https://doi.org/10.55776/DFH23}), and the PI project \href{https://doi.org/10.55776/P36413}{\textcolor{blue}{\textit{Intent-aware Music Recommender Systems}}} (\url{https://doi.org/10.55776/P36413}).
For open access purposes, the authors have applied a CC BY public copyright license to any author-accepted manuscript version arising from this submission.

\vfill\pagebreak


\bibliographystyle{IEEEbib}
\small
\bibliography{strings,refs}

\end{document}